\documentclass{article}
\usepackage{arxiv}

\usepackage{mathtools} 

\usepackage{booktabs}
\usepackage{array,multirow}
\usepackage{rotating}
\usepackage{makecell}
\usepackage{rotating}
\usepackage{array}
\usepackage{float}
\usepackage{physics}
\usepackage{algorithm}
\usepackage{algpseudocode}
\usepackage{enumitem}
\usepackage[font={small,it}]{caption}
\usepackage{hyperref}
\usepackage{microtype}

\usepackage{amssymb}
\usepackage{amsmath}
\usepackage{amsthm}
\usepackage{mathdots}
\usepackage{mathtools}

\usepackage{tensor}

\usepackage{urwchancal}
\DeclareFontFamily{OT1}{pzc}{}
\DeclareFontShape{OT1}{pzc}{m}{it}{<-> s * [1.10] pzcmi7t}{}
\DeclareMathAlphabet{\mathpzc}{OT1}{pzc}{m}{it}

\usepackage{graphicx}
\usepackage{ifpdf}
\ifpdf
\usepackage{epstopdf}
\PrependGraphicsExtensions{.svg}
\fi

\usepackage{xypic}

\graphicspath{{./images/}}
\newcommand{\etal}{\textit{et al}. }
\newcommand{\ie}{\textit{i}.\textit{e}.}
\newcommand{\eg}{\textit{e}.\textit{g}.}
\newcommand{\panelwidth}{1}

\renewcommand{\vec}[1]{\boldsymbol{#1}} %
\newcommand{\overbar}[1]{\mkern 1.5mu\overline{\mkern-1.5mu#1\mkern-1.5mu}\mkern 1.5mu}

\usepackage{wrapfig}
\newcommand{\eventframe}{event/frame }

\newcommand{\publicationdetails}
{Wang, Z., Ng, Y., Scheerlinck, C., Mahony R. (2021), ``An Asynchronous Kalman Filter for Hybrid Event Cameras", \textit{published in IEEE/CVF International Conference on Computer Vision (ICCV), 2021, pp. 448-457, IEEE DOI: \href{https://openaccess.thecvf.com/content/ICCV2021/html/Wang_An_Asynchronous_Kalman_Filter_for_Hybrid_Event_Cameras_ICCV_2021_paper.html}{10.1109/ICCV48922.2021.00050}} \\
	\copyrightNoticeIEEE{2021}	\\
}
\newcommand{\publicationversion}
{Author Accepted Version}

\begin{document}

\title{An Asynchronous Kalman Filter for Hybrid Event Cameras}
\headertitle{An Asynchronous Kalman Filter for Hybrid Event Cameras}

\author{
	\href{https://orcid.org/0000-0003-0815-1287}{\includegraphics[scale=0.06]{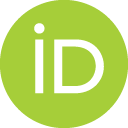}\hspace{1mm}
		Ziwei Wang}
	\\
	Systems Theory and Robotics Group \\
	Australian National University \\
	ACT, 2601, Australia \\
	\texttt{ziwei.wang1@anu.edu.au} \\
	\\
	\And
	\href{https://orcid.org/0000-0002-7764-298X}{\includegraphics[scale=0.06]{orcid.png}\hspace{1mm}
		Yonhon Ng}
	\\
	Systems Theory and Robotics Group \\
	Australian National University \\
	ACT, 2601, Australia \\
	\texttt{yonhon.ng@anu.edu.au} \\
	\\
	\And
	\href{https://orcid.org/0000-0003-0683-7963}{\includegraphics[scale=0.06]{orcid.png}\hspace{1mm}
		Cedric Scheerlinck}
	\\
	Systems Theory and Robotics Group \\
	Australian National University \\
	ACT, 2601, Australia \\
	\texttt{cedric.scheerlinck@anu.edu.au} \\
	\And	\href{https://orcid.org/0000-0002-7803-2868}{\includegraphics[scale=0.06]{orcid.png}\hspace{1mm}
		Robert Mahony}
	\\
	Systems Theory and Robotics Group \\
	Australian National University \\
	ACT, 2601, Australia \\
	\texttt{robert.mahony@anu.edu.au} \\
}

\maketitle

\begin{abstract}
Event cameras are ideally suited to capture HDR visual information without blur but perform poorly on static or slowly changing scenes.
Conversely, conventional image sensors measure absolute intensity of slowly changing scenes effectively but do poorly on high dynamic range or quickly changing scenes.
In this paper, we present an event-based video reconstruction pipeline for High Dynamic Range (HDR) scenarios.
The proposed algorithm includes a frame augmentation pre-processing step that deblurs and temporally interpolates frame data using events.
The augmented frame and event data are then fused using a novel asynchronous Kalman filter under a unifying uncertainty model for both sensors.
Our experimental results are evaluated on both publicly available datasets with challenging lighting conditions and fast motions and our new dataset with HDR reference.
The proposed algorithm outperforms state-of-the-art methods in both absolute intensity error (48\% reduction) and image similarity indexes (average 11\% improvement).
\end{abstract}

\centerline{
	\noindent \textbf{Code, Datasets and Video:}
}
\centerline{
	\noindent \href{https://github.com/ziweiWWANG/AKF}{https://github.com/ziweiWWANG/AKF}
}

\section{Introduction}
Event cameras offer distinct advantages over conventional frame-based cameras: high temporal resolution, high dynamic range (HDR) and minimal motion blur \cite{lichtsteiner2008128}.
However, event cameras provide poor imaging capability in slowly varying or static scenes, where despite some efforts in `gray-level' event cameras that measure absolute intensity \cite{posch2010qvga,Chen19cvprw}, most sensors predominantly measure only the relative intensity change.
Conventional imaging technology, conversely, is ideally suited to imaging static scenes and measuring absolute intensity.
Hybrid sensors such as the Dynamic and Active Pixel Vision Sensor (DAVIS) \cite{brandli2014240} or custom-built systems \cite{wang2020joint} combine event and frame-based cameras, and there is an established literature in video reconstruction fusing conventional and event camera data \cite{Scheerlinck18accv,Pan19cvpr, Pan20pami,wang2020joint}.
The potential of such algorithms to enhance conventional video to overcome motion blur and increase dynamic range has applications from robotic vision systems (\eg, autonomous driving), through film-making to smartphone applications for everyday use.

\begin{figure}
	\centering
	\resizebox{0.5\textwidth}{!}{
		\begin{tabular}{ c c }
			\includegraphics[width=0.49\linewidth]{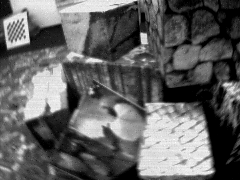} &
			\includegraphics[width=0.49\linewidth]{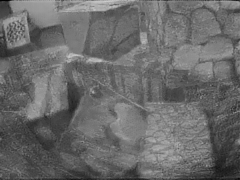}
			\\
			\LARGE (a) Input LDR Image & \LARGE (b) E2VID~\cite{Rebecq20pami}
			\\
			\\
			\includegraphics[width=0.49\linewidth]{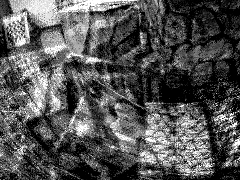} &
			\includegraphics[width=0.49\linewidth]{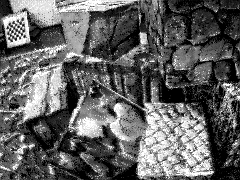}
			\\
			\LARGE (c) CF~\cite{Scheerlinck18accv}  & \LARGE (d) \textbf{Our AKF Reconstruction}
		\end{tabular}
	}
	\caption{
		An example with over exposure and fast camera motion causing blur taken from the open-source event camera dataset IJRR~\cite{Mueggler17ijrr}.
		Image (a) is the low dynamic range (LDR) and blurry input image.
		Image (b) is the result of state-of-the-art method E2VID~\cite{Rebecq20pami} (uses events only).
		Image (c) is the result of filter-based image reconstruction method CF~\cite{Scheerlinck18accv} that fuses events and frames.
		Our AKF (d) generates sharpest textured details in the overexposed areas.
	}
	\label{fig:front page}
\end{figure}

In this paper, we propose an Asynchronous Kalman Filter (AKF) to reconstruct HDR video from hybrid \eventframe cameras.
The key contribution is based on an explicit noise model we propose for both events and frames.
This model is exploited to provide a stochastic framework in which the pixel intensity estimation can be solved using an Extended Kalman Filter (EKF) algorithm \cite{kalman1960new,kalman1961new}.
By exploiting the temporal quantisation of the event stream, we propose an exact discretisation of the EKF equations, the Asynchronous Kalman Filter (AKF), that is computed only when events occur.
In addition, we propose a novel temporal interpolation scheme and apply the established de-blurring algorithm \cite{Pan20pami} to preprocess the data in a step called \emph{frame augmentation}.
The proposed algorithm demonstrates state-of-the-art hybrid event/frame image reconstruction as shown in Fig.~\ref{fig:front page}.

We compare our proposed algorithm with the state-of-the-art event-based video reconstruction methods on the popular public datasets ACD~\cite{Scheerlinck18accv}, CED~\cite{scheerlinck2019ced} and IJRR~\cite{Mueggler17ijrr} with challenging lighting conditions and fast motions.
However, existing public datasets using DAVIS event cameras do not provide HDR references for quantitative evaluation.
To overcome this limitation, we built a hybrid system consisting of a high quality RGB frame-based camera mounted alongside a pure event camera to collect high quality events, and HDR groundtruth from multiple exposures taken from the RGB camera.
Thus, we also evaluate the qualitative and quantitative performance of our proposed algorithm on our proposed HDR hybrid \eventframe dataset. Our AKF achieves superior performance to existing event and event/frame based image reconstruction algorithms.

In summary, our contributions are:
\begin{itemize}[noitemsep]
	\item An Asynchronous Kalman Filter (AKF) for hybrid \eventframe HDR video reconstruction
	\item A unifying \eventframe uncertainty model
	\item Deblur and temporal interpolation for frame augmentation
	\item A novel real-world HDR hybrid \eventframe dataset with reference HDR images and a simulated HDR dataset for quantitative evaluation of HDR performance.
\end{itemize}

\section{Related Work}
Recognising the limited ability of pure event cameras (DVS) \cite{lichtsteiner2008128} to detect slow/static scenes and absolute brightness, hybrid \eventframe cameras such as the DAVIS \cite{brandli2014240} were developed.
Image frames and events are captured through the same photodiode allowing the two complementary data streams to be exactly registered \cite{Brandli14iscas}.
This has led to significant research effort into image reconstruction from hybrid \eventframe and pure event cameras including SLAM-based methods \cite{Kim16eccv,Rebecq17ral}, filters \cite{Scheerlinck18accv,Scheerlinck19ral}, de-blurring \cite{Pan19cvpr,Pan20pami}, machine learning approaches \cite{Rebecq20pami,Scheerlinck20wacv,Stoffregen20eccv}.

Video and image reconstruction methods may be grouped into (i) per-event asynchronous algorithms that process events upon arrival \cite{Brandli14iscas,Wang19acra,Scheerlinck18accv} and (ii) batch (synchronous) algorithms that first accumulate a significant number (\eg, 10k) of events before processing the batch in one go \cite{Pini19iciap, Rebecq20pami,Scheerlinck20wacv}.
While batch methods have achieved high accuracy, they incur additional latency depending on the time-interval of the batch (\eg, 50ms).
Asynchronous methods, if implemented on appropriate hardware, have the potential to run on a timescale closer to that of events $< 1$ms.
A further distinction may be made between pure event reconstruction methods and hybrid \eventframe methods that use a mix of (registered) events and image frames.

\noindent
\textbf{Pure event reconstruction:}
Images and video reconstruction using only events is a topic of significant interest in the community that can shed light on the information content of events alone.
Early work focused on a moving event camera in a static scene, either pure rotations \cite{Cook11ijcnn,Kim14bmvc} or full 6-DOF motion \cite{Kim16eccv,Rebecq17ral}.
Hand-crafted approaches were proposed including joint optimisation over optic flow and image intensity \cite{Bardow16cvpr}, periodic regularisation based on event timestamps \cite{Reinbacher16bmvc} and temporal filtering \cite{Scheerlinck18accv,Scheerlinck19ral}.
Recently, learned approaches have achieved surprisingly high quality video reconstruction \cite{Rebecq19cvpr,Rebecq20pami,Scheerlinck20wacv,Stoffregen20eccv} at significantly higher computational cost vs. hand-crafted methods.

\noindent
\textbf{Event/frame reconstruction:}
The invention of the DAVIS \cite{brandli2014240} and its ability to capture frames alongside events (and even IMU measurements) has widened the community's perspective from pure event cameras to hybrid sensors and how best to combine modalities.
An early algorithm interpolated between frames by adding events scaled by the contrast threshold until a new frame is received  \cite{Brandli14iscas}.
The contrast threshold is typically unknown and variable so \cite{Brandli14iscas} includes a method to estimate it based on surrounding image frames from the DAVIS.
Pan \etal \cite{Pan19cvpr,Pan20pami} devised the event double integral (EDI) relation between events and a blurry image, along with an optimisation approach to estimate contrast thresholds to reconstruct high-speed de-blurred video from events and frames.
High-speed video can also be obtained by warping still images according to motion computed via events \cite{Shedligeri19jei,Liu17tvc}, or by letting a neural network learn how to combine frames and events \cite{Pini20visigrapp,Wang19iccvw,Pini19iciap,Lin20eccv,Jiang20cvpr}.
Recognising the limited spatial resolution of the DAVIS, Han \etal \cite{han2020neuromorphic} built a hybrid event/frame system consisting of an RGB camera and a DAVIS240 event camera registered via a beam-splitter.
An event guided HDR imaging pipeline was used to fuse frame and event information \cite{han2020neuromorphic}.

Continuous-time temporal filtering is an approach that exploits the near-continuous nature of events.
Scheerlinck \etal~\cite{Scheerlinck18accv,Scheerlinck19ral} proposed an asynchronous complementary filter to fuse events and frames that can equivalently be run as a high-pass filter if the frame input is set to zero (\ie, using events only).
The filters are based on temporal smoothing via a single fixed-gain parameter that determines the `fade rate' of the event signal.

\noindent
\textbf{Multi-exposure image fusion (MEIF):}
The most common approach in the literature to compute HDR images is to fuse  multiple images taken with different exposures.
Ma~\emph{et al.}~\cite{Ma17TIP} proposed the use of structural patch decomposition to handle dynamic objects in the scene.
Kalantari and Ramamoorthi~\cite{Kalantari17ACMTG} proposed a deep neural network and a dataset for dynamic HDR MEIF.
More recent work also deals with motion blur in long exposure images~\cite{Wang18ICSIP, Li18ICIP}.
These methods directly compute images that do not require additional tone mapping to produce nice looking images~\cite{Que19TIM}.
However, all these works require multiple images at different exposures of the same scene and cannot be applied to the real-time image reconstruction scenarios considered in this paper.

\section{Sensor Model and Uncertainty}
\subsection{Event Camera Model}\label{sec:Event Camera Mode}
Event cameras measure the relative log intensity change of irradiance of pixels.
New events $e_{\vec{p}}^i$ are triggered when the log intensity change exceeds a preset contrast threshold $c$.
In this work, we model events as a Dirac delta or impulse function $\delta$ \cite{astrom2010feedback} to allow us to apply continuous-time systems analysis for filter design.
That is,
\begin{align}
e_{\vec{p}}(t) &=
\sum_{i=1}^\infty
(c \sigma_{\vec{p}}^i + \eta^i_{\vec{p}}) \delta(t-t^i_{\vec{p}}), \label{eq:continuous_event}
\\
\eta^i_{\vec{p}} &\sim
\mathcal{N}\left(
0, Q_{\vec{p}}(t)
\right), \notag
\end{align}
where $t^i_{\vec{p}}$ is the time of the $i^{th}$ event at the $\vec{p} = (\vec{p}_x ,\vec{p}_y)^{T}$ pixel coordinate, the polarity $\sigma^i_{\vec{p}} \in \{-1,+1\}$ represents the direction of the log intensity change, and the noise $\eta^i_{\vec{p}}$ is an additive Gaussian uncertainty at the instance when the event occurs.
The noise covariance $Q_{\vec{p}}(t)$ is the sum of three contributing noise processes; `process' noise, `isolated pixel' noise, and `refractory period' noise.
That is
\begin{align}
\label{eq:event_covariance}
Q_{\vec{p}}(t) :=
\sum_{i = 1}^{\infty}\left(Q^\text{proc.}_{\vec{p}}(t) +
Q^\text{iso.}_{\vec{p}}(t) + Q^\text{ref.}_{\vec{p}}(t)
\right)
\delta(t - t^i_{\vec{p}}).
\end{align}
We further discuss the three noise processes in the next section.

\subsubsection{Event Camera Uncertainty}\label{sec:Q_model}
Stochastic models for event camera uncertainty are difficult to develop and justify~\cite{gallego2019event}.
In this paper, we propose a number of simple heuristics to model event noise as the sum of three pixel-by-pixel additive Gaussian processes.

\noindent
\textbf{Process noise:}
Process noise is a constant additive uncertainty in the evolution of the irradiance of the pixel, analogous to process noise in a Kalman filtering model.
Since this noise is realised as an additive uncertainty only when an event occurs, we call on the principles of Brownian motion to model the uncertainty at time $t^i_{\vec{p}}$ as a Gaussian process with covariance that grows linearly with time since the last event at the same pixel.
That is
\begin{align*}
Q^\text{proc.}_{\vec{p}}(t^i_{\vec{p}})
& = \sigma_{\text{proc.}}^2 (t^i_{\vec{p}} - t^{i-1}_{\vec{p}}),
\end{align*}
where $\sigma_{\text{proc.}}^2$ is a tuning parameter associated with the process noise level.

\noindent
\textbf{Isolated pixel noise:}
Spatially and temporally isolated events are more likely to be associated to noise than events that are correlated in group.
The noisy background activity filter~\cite{delbruck2008frame} is designed to suppress such noise and most event cameras have similar routines that can be activated.
Instead, we model an associated noise covariance by
\[
Q^\text{iso.}_{\vec{p}}(t^i_{\vec{p}})
= \sigma_{\text{iso.}}^2 \min \{t^i_{\vec{p}} - t^*_{N(\vec{p})}\},
\]
where $\sigma_{\text{iso.}}^2$ is a tuning parameter and $t^*_{N(\vec{p})}$ is the latest time-stamp of any event in a neighbourhood $N(\vec{p})$ of $\vec{p}$.
If there are recent spatio-temporally correlated events then $Q^\text{iso.}_{\vec{p}}(t^i_{\vec{p}}) $ is negligible, however, the covariance grows linearly, similar to the Brownian motion assumption for the process noise, with time from the most recent event.

\noindent
\textbf{Refractory period noise:}
Circuit limitations in each pixel of an event camera limit the response time of events to a minimum known as the refractory period $\rho > 0$ \cite{yang2015dynamic}.
If the event camera experience fast motion in highly textured scenes then the pixel will not be able to trigger fast enough and events will be lost.
We model this by introducing a dependence on the uncertainty associated with events that are temporally correlated such that
\[
Q^\text{ref.}_{\vec{p}}(t^i_{\vec{p}}) =
\left\{
\begin{array}{cc}
0 & \text{if } t^i_{\vec{p}} - t^{i-1}_{\vec{p}} > \overbar{\rho}, \\
\sigma_{\text{ref.}}^2 & \text{otherwise} ,
\end{array}
\right.
\]
where $\sigma_{\text{ref.}}^2$ is a tuning parameter and $\overbar{\rho}$ is an upper bound on the refractory period.

\subsection{Conventional Camera Model}
\label{sec:Conventional Camera Model}
The photo-receptor in a CCD or CMOS circuit from a conventional camera converts incoming photons into charge that is then converted to a pixel intensity by an analogue-to-digital converter (ADC).
In a typical camera, the camera response is linearly related to the pixel irradiance for the correct choice of exposure, but can become highly non-linear where pixels are overexposed or underexposed \cite{madden1993extended}.
In particular, effects such as dark current noise, CCD saturation, and blooming destroy the linearity of the camera response at the extreme intensities \cite{kim2008characterization}.
In practice, these extreme values are usually trimmed, since the data is corrupted by sensor noise and quantisation error.
However, the information that can be gained from this data is critically important for HDR reconstruction.
The mapping of the scaled sensor irradiance (a function of scene radiance and exposure time) to the camera response is termed the Camera Response Function (CRF) \cite{grossberg2003space,robertson2003estimation}.
To reconstruct the scaled irradiance $I_{\vec{p}}(\tau^k)$ at pixel $\vec{p}$ at time $\tau^k$ from the corresponding raw camera response $I^F_{\vec{p}}(\tau^k)$ one applies the inverse CRF
\begin{align}
I_{\vec{p}}(\tau^k) = CRF^{-1}( I^F_{\vec{p}}(\tau^k) )  + \bar{\mu}_{\vec{p}}^k, \label{eq:I_F}
\\
\bar{\mu}_{\vec{p}}^k \sim \mathcal{N}(0,\bar{R}_{\vec{p}}(\tau^k)), \notag
\end{align}
where $\bar{\mu}_{\vec{p}}^k$ is a noise process that models noise in $I_{\vec{p}}(\tau^k)$ corresponding to noise in $I^F_{\vec{p}}$ mapped back through the inverse CRF.
This inverse mapping of the noise is critical in correctly modelling the uncertainty of extreme values of the camera response.
%

\subsubsection{Conventional Camera Uncertainty}
The noise of $I_{\vec{p}}(\tau^k)$ comes from uncertainty in the raw camera response $I^F_{\vec{p}}(\tau^k)$ mapped through the inverse of the Camera Response Function (CRF).
The uncertainty associated with sensing process $I^F_{\vec{p}}(\tau^k)$ is usually modelled as a constant variance Gaussian process
\cite{shin2005block, russo2003method} although for low light situations this should properly be a Poisson process model \cite{hasinoff2014photon}.
The quantisation noise is uniform over the quantisation interval related to the number of bits used for intensity encoding.
Since the CRF compresses the sensor response for extreme intensity values, the quantisation noise will dominate in these situations.
Conversely, for correct exposure settings, the quantisation noise is insignificant and a Gaussian sensing process uncertainty provides a good model \cite{hasinoff2014photon}.
Inverting this noise model through the inverse of the CRF function then we expect the variance $\bar{R}_{\vec{p}}(\tau^k)$ in \eqref{eq:I_F} to depend on intensity of the pixel: it should be large for extreme intensity values and roughly constant and small for well exposed pixels.

The CRF can be estimated using an image sequence taken under different exposures~\cite{debevec2008recovering,grossberg2003space,robertson2003estimation}.
For long exposures, pixels that would have been correctly exposed become overexposed and provide information on the nonlinearity of the CRF at high intensity, and similarly, short exposures provide information for the low intensity part of the CRF. We have used this approach to estimate the CRF for the APS sensor on a DAVIS event camera and a FLIR camera.
In the experiment, we use the raw image intensity as the measured camera response.

Following \cite{robertson2003estimation}, the exposure time is linearly scaled to obtain the scaled irradiance in the range of raw camera response.
In this way, the camera response function $\text{CRF}(\cdot)$ is experimentally determined as a function of the scaled irradiance $I$.
The Certainty function $f^{c} (\cdot)$ is defined to be the sensitivity of the CRF with respect to the scaled irradiance
\begin{align}
\label{eq:certainty function}
f^{c} & :=
\frac{
	\mathrm{d} \text{CRF}
}{
\mathrm{d} I
},
\end{align}
and it is renormalised so that the maximum is unity \cite{robertson2003estimation}.
Note that different cameras can have dissimilar camera responses for the same irradiance of the sensor.

\begin{figure}[t!]
	\centering
	\begin{tabular}{c c}
		\\
		\includegraphics[width=0.33\linewidth]{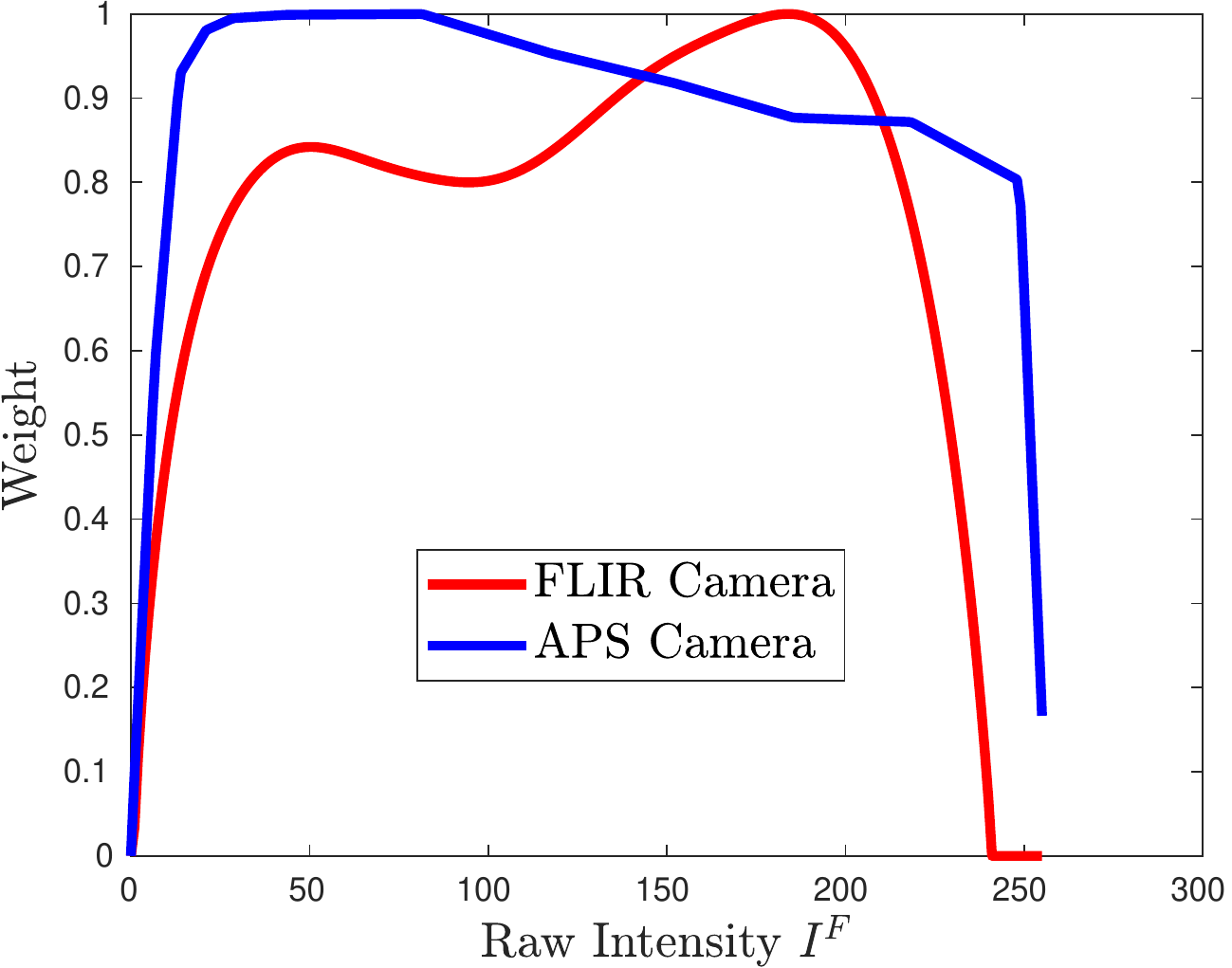} &
		\includegraphics[width=0.33\linewidth]{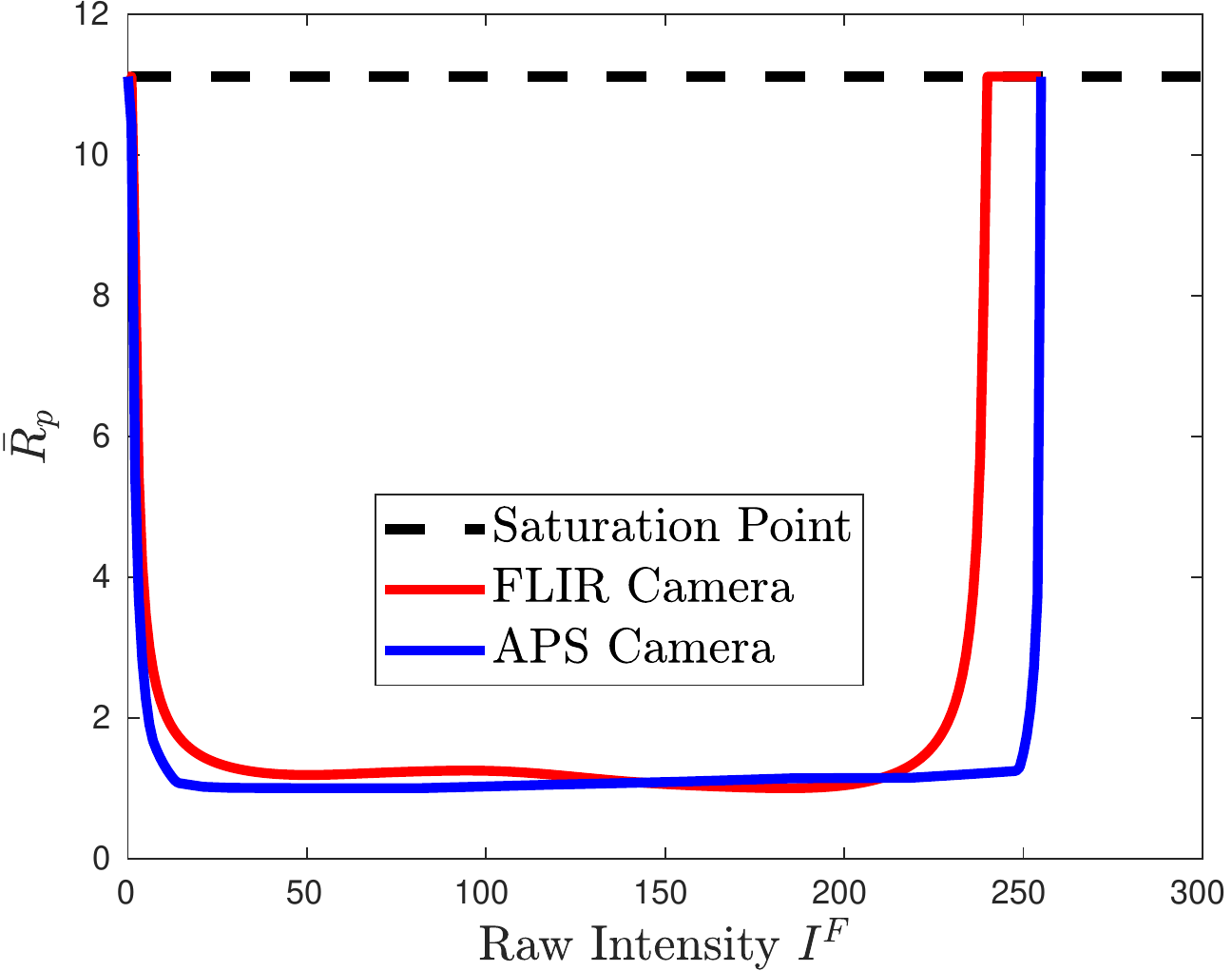}
		\\
		(a) Weighting function  & (b) Image covariance function
		\\
	\end{tabular}	
	\caption{\label{fig:crf}
		Weighting function $f^w(\cdot)$ and image covariance function $\bar{R}_{\vec{p}}$ for the APS camera in a DAVIS \eventframe camera (blue) and the \textit{FLIR} camera (red) used in the experimental studies.
	}
\end{figure}

Remapping the $I$ axis of the Certainty function $f^c(\cdot)$ to camera response $I^F$ defines the Weighting function $f^{w} (\cdot)$ (Fig~\ref{fig:crf}.a) as a function of camera response \cite{robertson2003estimation}
\begin{align}
\label{eq:weight function}
f^{w} & :=
\frac{
	\mathrm{d} \text{CRF}
}{
\mathrm{d} I
} \circ \text{CRF}^{-1},
\end{align}
where $\circ$ defines function composition.

Inspired by \cite{robertson2003estimation}, we define the covariance of noise associated with raw camera response as
\begin{align}
\label{eq:sigma_i}
\bar{R}_{\vec{p}} & :=  {\sigma}_{\text{im.}}^2
\frac{1}{f^{w}(I^F)},
\end{align}
where $\sigma_{\text{im.}}^2$ is a tuning parameter related to the base level of noise in the image
(see Fig. \ref{fig:crf}.b. for $\sigma_{\text{im.}}^2 = 1$).
Note that we also introduce a saturation to assign a maximum value to the image covariance function (Fig.~\ref{fig:crf}.b).

In addition to the base uncertainty model for $I_{\vec{p}}(\tau^{k})$, we will also need to model the uncertainty of frame information in the interframe period and in the log intensity scale for the proposed algorithm.
We use linear interpolation to extend the covariance estimate from two consecutive frames $I_{\vec{p}}(\tau^{k})$ and $I_{\vec{p}}(\tau^{k+1})$ by
\begin{align}
\label{eq: compute R 1}
\bar{R}_{\vec{p}}(t) := \Big(\frac{t-\tau^{k}}{\tau^{k+1} - \tau^{k}}\Big) \bar{R}_{\vec{p}}(\tau^{k+1}) + \Big(\frac{\tau^{k+1}-t}{\tau^{k+1} - \tau^{k}}\Big) \bar{R}_{\vec{p}}(\tau^{k}).
\end{align}
We define the continuous log image intensity function by taking the log of $I_{\vec{p}}$.
However, the log function is not symmetric and mapping the noise from $I_{\vec{p}}$ will bias the log intensity.
Using Taylor series expansion, the biased log intensity is approximately
\begin{align}
L_{\vec{p}}^F(\tau^k) \approx \log(I_{\vec{p}}(\tau^k) + I_0) - \frac{\bar{R}_{\vec{p}}(\tau^k)}{2(I_{\vec{p}}(\tau^k) + I_0)^2} + \mu_{\vec{p}}^k,  \notag \\
\mu_{\vec{p}}^k \sim \mathcal{N}(0,R_{\vec{p}}(\tau^k)),
\label{eq:L_F}
\end{align}
where $I_0$ is a fixed offset introduced to ensure intensity values remain positive and
$R_{\vec{p}}(\tau^k)$ is the covariance of noise associated with the log intensity.
The covariance is given by
\begin{align}
\label{eq:frame_covariance}
R_{\vec{p}}(t) &= \frac{\bar{R}_{\vec{p}}(t)}{(I_{\vec{p}}(\tau^k) + I_0)^2}.
\end{align}
Generally, when $I_{\vec{p}}(\tau^k)$ is not extreme then
$\frac{\bar{R}_{\vec{p}}(t)}{2(I_{\vec{p}}(\tau^k) + I_0)^2} \ll \log(I_{\vec{p}}(\tau^k) + I_0)$ and
$L_{\vec{p}}^F(\tau^k) \approx \log(I_{\vec{p}}(\tau^k) + I_0)$.

\section{Method}\label{sec:method}

\begin{figure*}
	\centering
	\includegraphics[trim=3 5 2 5, clip, width=0.95\linewidth]{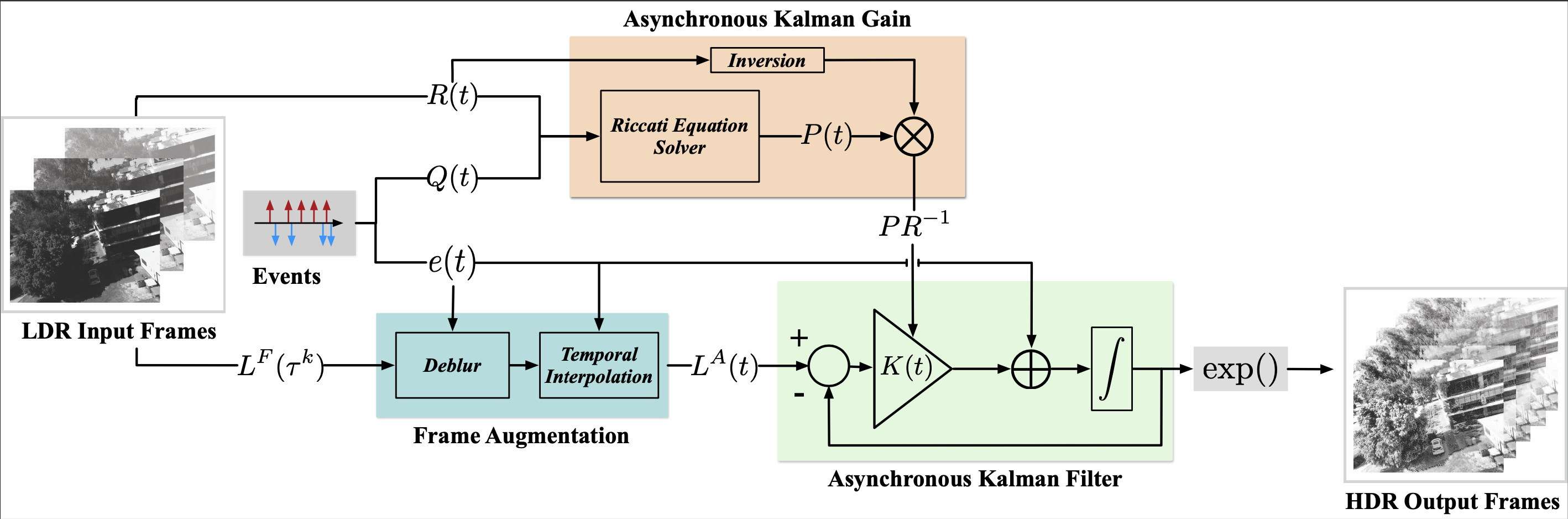}
	\caption{
		Block diagram of the image processing pipeline discussed in \S\ref{sec:method}.}
	\label{fig: pipeline}
\end{figure*}

The proposed image processing architecture is shown in Fig. \ref{fig: pipeline}.
There are three modules in the proposed algorithm; a frame augmentation module that uses events to augment the raw frame data to remove blur and increase temporal resolution, the Asynchronous Kalman Filter (AKF) that fuses the augmented frame data with the event stream to generate HDR video, and the Kalman gain module that integrates the uncertainty models to compute the filter gain.

\subsection{Frame Augmentation}
\begin{figure}
	\centering
	\includegraphics[width=0.55\linewidth]{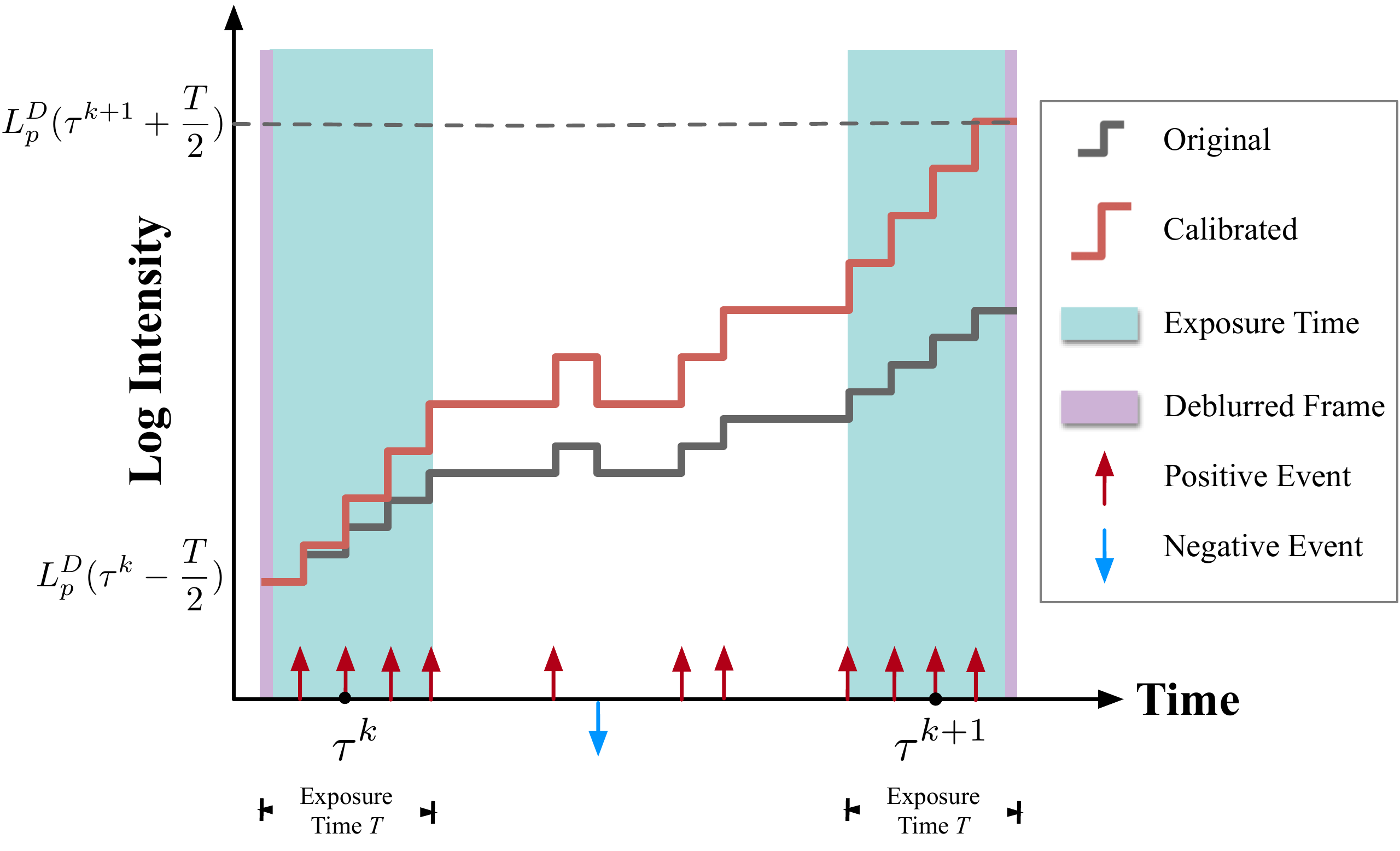}
	\caption{\label{fig: preprocessing}
		Frame augmentation.
		Two deblurred frames at times $\tau^k - \frac{T}{2}$ and $\tau^{k+1} + \frac{T}{2}$ are computed.
		The event stream is used to interpolate between the two deblurred frames to improve temporal resolution.
	} 
\end{figure}

\noindent\textbf{Deblur:}
Due to long exposure time or fast motion, the intensity images $L^F$ may suffer from severe motion blur.
We use the double integral model (EDI) from~\cite{Pan19cvpr} to sharpen the blurry low frequency images to obtain a deblurred image $L_{\vec{p}}^D(\tau^k - T/2)$ at the beginning, and
$L_{\vec{p}}^D(\tau^{k+1} + T/2)$ at the end, of the exposure of each frame (Fig.~\ref{fig: preprocessing}).
The two sharpened images are used in the interpolation module.

\noindent\textbf{Interpolation:}
The goal of the interpolation module is to increase the temporal resolution of the frame data.
This is important to temporally align the information in the image frames and event data, which helps to overcome the ghosting effects that are visible in other recent work where the image frames are interpolated using zero order hold \cite{Scheerlinck18accv, Scheerlinck19ral}.

To estimate intensity at the $i^{th}$ event timestamp at pixel $\vec{p}$, we integrate forward from a deblurred image $L_{\vec{p}}^D (\tau^k - T/2)$ taken from the start of the exposure (Fig.~\ref{fig: preprocessing}).
The forward interpolation is
\begin{equation} \label{eq:interpolation1}
L_{\vec{p}}^{A-}(t) = L_{\vec{p}}^D(\tau^{k} - T/2) + \int_{\tau^{k} - T/2}^{t} e(\gamma) d\gamma,
\end{equation}
where $L_{\vec{p}}^{A-}$ denotes the augmented image.
Similarly, we interpolate backwards from the end of exposure $k+1$ to obtain
\begin{equation} \label{eq:interpolation2}
L_{\vec{p}}^{A+}(t) = L_{\vec{p}}^D(\tau^{k+1} + T/2) - \int_{t}^{\tau^{k+1} + T/2} e(\gamma) d\gamma.
\end{equation}

Ideally, if there are no missing or biased events and the frame data is not noisy, then the forwards and backwards interpolation results $L_{\vec{p}}^{A-}(t^{i}_{\vec{p}})$ and $L_{\vec{p}}^{A+}(t^{i}_{\vec{p}})$ computed with the true contrast threshold should be equal.
However, noise in either the event stream or in the frame data will cause the two interpolations to differ.
We reconcile these two estimates by per-pixel calibration of the contrast threshold in each interpolation period.
Define the scaling factor of the contrast threshold
\begin{equation} \label{eq:scaling factor}
c^k_{\vec{p}} := \frac{L_{\vec{p}}^D(\tau^{k+1} + T/2) - L_{\vec{p}}^D(\tau^{k} - T/2)}{\int_{\tau^{k} - T/2}^{\tau^{k+1} + T/2} e(\gamma) d\gamma}.
\end{equation}
This calibration can be seen as using the shape provided by the event integration between deblurred frames and scaling the contrast threshold to vertically stretch or shrink the interpolation to fit the deblurred frame data (Fig.~\ref{fig: preprocessing}).
This is particularly effective at compensating for refractory noise where missing events are temporally correlated to the remaining events.
Using the outer limits of the exposure for the deblurred image maximises the number of events (per-pixel) in the interpolation period and improves the estimation of $c^k_{\vec{p}}$.

Within each exposure (frame $k$) there is a forward and backward estimate available with different per-pixel contrast thresholds associated with interpolating from frame $k-1$ to $k$, $k$ to $k+1$.
We smoothly interpolate between estimates in the exposure period to define the final augmented frame
\begin{equation}
	L_{\vec{p}}^A(t) =
	\left\{
	\begin{array}{l}
	\Bigg(\frac{\tau^{k} + T/2  - t}{T}\Bigg) L_{\vec{p}}^{A-}(t)
	+ \Bigg(\frac{t-\tau^{k} + T/2}{T}\Bigg) L_{\vec{p}}^{A+}(t) \\
	\quad\quad\quad\quad\quad\quad \text{ if } t \in [\tau^k -T/2, \tau^k + T/2), \\
	L_{\vec{p}}^{A+}(t) \\
	\quad\quad\quad\quad\quad \text{ if }  t \in [\tau^k +T/2, \tau^{k+1} - T/2).
	\end{array}
	\right.
\label{eq:interpolation3}
\end{equation}

\subsection{Asynchronous Kalman Filter (AKF)}
In this section, we introduce the Kalman filter that integrates the uncertainty models of both event and frame data to compute the filter gain dynamically.
We propose a continuous-time stochastic model of the log intensity state
\begin{align*}
\mathrm{d} L_{\vec{p}}
&= e_{\vec{p}}(t)\mathrm{d}t + \mathrm{d} w_{\vec{p}}, \\
L_{\vec{p}}^{A}(t^{i}_{\vec{p}})  &=
L_{\vec{p}}(t^{i}_{\vec{p}}) + \mu_{\vec{p}}^i,
\end{align*}
where $\mathrm{d} w_{\vec{p}}$ is a Wiener process (continuous-time stochastic process) and $\mu_{\vec{p}}^i$ is the log intensity frame noise \eqref{eq:L_F} in continuous time associated with the models introduced in \S\ref{sec:Event Camera Mode} and \S\ref{sec:Conventional Camera Model}.
Here $L^A_{\vec{p}}(t^{i}_{\vec{p}})$ is the augmented image (see $L^A (t)$ in Fig.~\ref{fig: pipeline})
and the notation serves also as the measurement equation where $L_{\vec{p}}(t^{i}_{\vec{p}})$ is the true (log) image intensity.

The ordinary differential equation (ODE) of the proposed filter state estimate is
\begin{align}
\dot{\hat{L}}_{\vec{p}}(t) &= e_{\vec{p}}(t) -K_{\vec{p}}(t)[\hat{L}_{\vec{p}}(t) - L_{\vec{p}}^A(t)]
\label{eq:filter_ODE},
\end{align}
where $K_{\vec{p}}(t)$ is the Kalman gain defined below \eqref{eq:kf}.
The Kalman-Bucy filter that we implement is posed in continuous-time and updated asynchronously as each event arrives.
At each new event timestamp $t^i_{\vec{p}}$, the filter state is updated as
\begin{align}
\hat{L}_{\vec{p}}(t^{i}_{\vec{p}}) = \hat{L}_{\vec{p}}(t^{i-}_{\vec{p}}) + e_{\vec{p}}(t^i_{\vec{p}}). \label{eq:filter_e}
\end{align}
Within a time-interval \mbox{$t \in [t^i_{\vec{p}}, t^{i+1}_{\vec{p}})$} where there are no new events or frames we solve the following ODE as a discrete update
\begin{align}
\dot{\hat{L}}_{\vec{p}}(t) = -K_{\vec{p}}(t)[\hat{L}_{\vec{p}}(t) - L_{\vec{p}}^A(t)] \text{ for } t \in [t^i_{\vec{p}}, t^{i+1}_{\vec{p}}) \label{eq:filter_ODE_TimeInt}.
\end{align}
Substituting the Kalman gain $K_{\vec{p}}(t)$ from \eqref{eq:kf} and \eqref{eq:P_full}, the analytic solution of \eqref{eq:filter_ODE_TimeInt} between frames or events is
\begin{align} \label{eq:L_hat_full}
\hat{L}_{\vec{p}}(t) = \frac{ [\hat{L}_{\vec{p}}(t^{i}_{\vec{p}}) - L_{\vec{p}}^A(t^{i}_{\vec{p}})] \cdot P_{\vec{p}}^{-1}(t^{i}_{\vec{p}})}{P_{\vec{p}}^{-1}(t^{i}_{\vec{p}}) + R_{\vec{p}}^{-1}(t) \cdot  (t - t^{i}_{\vec{p}})} + L_{\vec{p}}^A(t).
\end{align}
The detailed derivation of $\hat{L}_{\vec{p}}(t)$ is shown in the supplementary material \S 6.

\begin{figure*}
	\centering
	\resizebox{1\textwidth}{!}{
		\begin{tabular}{
				>{\centering\arraybackslash}m{4mm}
				>{\centering\arraybackslash}m{6cm}
				>{\centering\arraybackslash}m{6cm} >{\centering\arraybackslash}m{6cm}
				>{\centering\arraybackslash}m{6cm}}
			\rotatebox{90}{\Large \texttt{Night drive}}
			&
			\includegraphics[width=\panelwidth\linewidth]{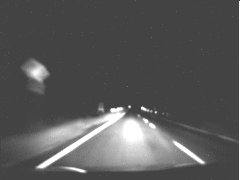}
			&
			\includegraphics[width=\panelwidth\linewidth]{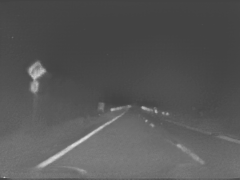}
			&
			\includegraphics[width=\panelwidth\linewidth]{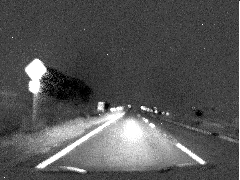}
			&
			\includegraphics[width=\panelwidth\linewidth]{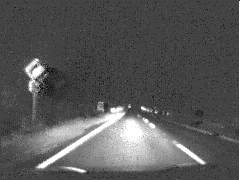}\\
			\rotatebox{90}{\Large \texttt{Shadow}}
			&
			\includegraphics[width=\panelwidth\linewidth]{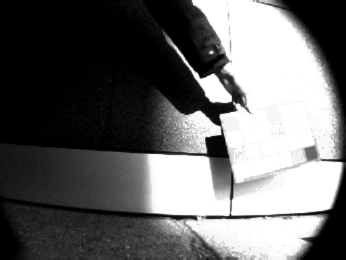}
			&
			\includegraphics[width=\panelwidth\linewidth]{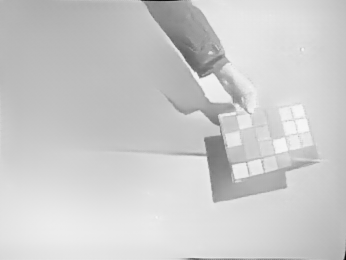}
			&
			\includegraphics[width=\panelwidth\linewidth]{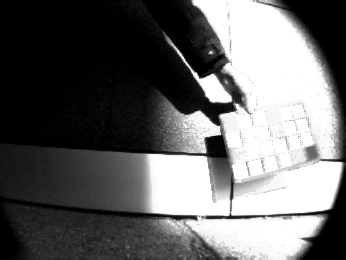}
			&
			\includegraphics[width=\panelwidth\linewidth]{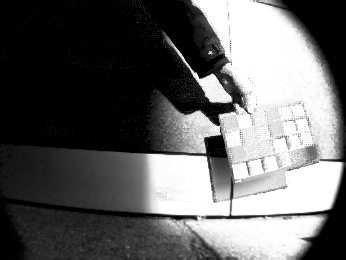}
			\\
			\rotatebox{90}{\Large \texttt{Outdoor running}}
			&	
			\includegraphics[width=\panelwidth\linewidth]{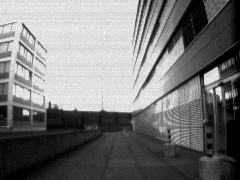}
			&
			\includegraphics[width=\panelwidth\linewidth]{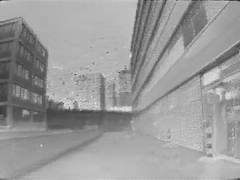}
			&
			\includegraphics[width=\panelwidth\linewidth]{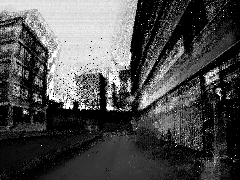}
			&
			\includegraphics[width=\panelwidth\linewidth]{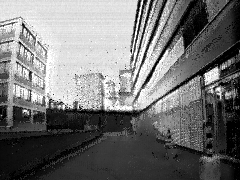}
			\\
			& \Large (a) DAVIS frame
			& \Large (b) E2VID \cite{Rebecq20pami}
			& \Large (c) CF \cite{Scheerlinck18accv}
			& \Large (d) \textbf{AKF(ours)}
		\end{tabular}
	}
	\caption{Comparison of state-of-the-art event-based video reconstruction methods on sequences with challenging lighting conditions and fast motions, drawn from the open-source datasets ACD~\cite{Scheerlinck18accv}, CED~\cite{scheerlinck2019ced} and IJRR~\cite{Mueggler17ijrr}.
		CF \cite{Scheerlinck18accv} fails to capture details under extreme lighting conditions and suffers from a `shadowing effect' (white or black shadows trailing behind dark or bright moving objects).
		E2VID \cite{Rebecq20pami} and AKF are able to reconstruct the blurry right turn sign in the high-speed, low-light \texttt{Night drive} dataset and the overexposed regions in the \texttt{Shadow} and \texttt{Outdoor running} dataset. 
		But without frame information, E2VID~\cite{Rebecq20pami} fails to compute the static background of \texttt{Shadow}, and only provides washed-out reconstructions in all three sequences.
		AKF outperforms the other methods in all challenging scenarios.
		Additional image and video comparisons are provided in the supplementary material.
	}
	\label{fig:davis}
\end{figure*}

\subsection{Asynchronous Kalman Gain}\label{sec:KalmanGain}
The Asynchronous Kalman filter computes a pixel-by-pixel gain $K_{\vec{p}}(t)$ derived from estimates of the state and sensor uncertainties.
The Kalman gain is given by \cite{kalman1960new,kalman1961new}
\begin{align}
\label{eq:kf}
K_{\vec{p}}(t) = P_{\vec{p}}(t) R_{\vec{p}}^{-1}(t),
\end{align}	
where $P_{\vec{p}}(t) > 0$ denotes the covariance of the state estimate in the filter and $R_{\vec{p}}(t)$ \eqref{eq:frame_covariance} is the log-intensity frame covariance of pixel $\vec{p}$.
The standard Riccati equation \cite{kalman1960contributions,zaitsev2002handbook} that governs the evolution of the filter state covariance~\cite{kalman1961new} is given by
\[
\dot{P}_{\vec{p}}(t) = -P_{\vec{p}}^2 R_{\vec{p}}^{-1}(t) + Q_{\vec{p}}(t),
\]
where $Q_{\vec{p}}(t)$ \eqref{eq:event_covariance} is the event noise covariance.
Here the choice of event noise model \eqref{eq:event_covariance} as a discrete noise that occurs when the update of information occurs means that the Riccati equation can also be solved during the time interval $t \in [t^i_{\vec{p}}, t^{i+1}_{\vec{p}})$ and at new event timestamp $t^{i+1}_{\vec{p}}$ separately.

In the time interval $t \in [t^i_{\vec{p}}, t^{i+1}_{\vec{p}})$ (no new events or frames occur), the
state covariance $P_{\vec{p}}(t)$ is asynchronously updated by the ordinary differential equation
\begin{align}
\dot{P}_{\vec{p}}(t) = - P_{\vec{p}}^2(t) \cdot R_{\vec{p}}^{-1}(t). \label{eq:Riccati_ODE1}
\end{align}
Computing $R_{\vec{p}}(t)$ from \eqref{eq:sigma_i}-\eqref{eq:frame_covariance} on this time interval then the solution of \eqref{eq:Riccati_ODE1} is
\begin{align}
P_{\vec{p}}(t) &= \frac{1}{P_{\vec{p}}^{-1}(t^{i}_{\vec{p}}) + R_{\vec{p}}^{-1}(t) \cdot  (t - t^{i}_{\vec{p}})}, \notag \\
& \quad\quad\quad\quad\quad\quad \text{ for }
t \in [t^i_{\vec{p}}, t^{i+1}_{\vec{p}}).
\label{eq:P_full}
\end{align}
At the new event timestamp $t^{i+1}_{\vec{p}}$, the
state covariance $P_{\vec{p}}(t)$ is updated from the timestamp $t^{{(i+1)}-}_{\vec{p}}$ such that
\begin{align}
P_{\vec{p}}(t^{{i+1}}_{\vec{p}}) & = P_{\vec{p}}(t^{{(i+1)}-}_{\vec{p}}) + Q_{\vec{p}}(t^{i+1}_{\vec{p}}). \label{eq:P_event}
\end{align}
The explicit solution of Kalman filter gain is obtained by substituting \eqref{eq:P_full} and \eqref{eq:P_event} to \eqref{eq:kf}. See derivation of $P_{\vec{p}}(t)$ in the supplementary material \S5.
The solution is substituted into \eqref{eq:filter_ODE} to obtain \eqref{eq:L_hat_full}.

\section{Hybrid Event/Frame Dataset}
Evaluating HDR reconstruction for hybrid \eventframe cameras requires a dataset including synchronised events, low dynamic range video and high dynamic range reference images.
The dataset associated with the recent work by~\cite{han2020neuromorphic} is patent protected and not publicly available.
Published datasets lack high quality HDR reference images, and instead rely on low dynamic range sensors such as the APS component of a DAVIS for groundtruth \cite{Stoffregen20eccv,Zhu18ral,Mueggler17ijrr}.
Furthermore, these datasets do not specifically target HDR scenarios.
DAVIS cameras used in these datasets also suffer from shutter noise (noise events triggered by APS frame readout) due to undesirable coupling between APS and DVS components of pixel circuitry \cite{brandli2014240}.

To address these limitations, we built a hybrid \eventframe camera system
consisting of two separate high quality sensors, a Prophesee event camera (VGA, 640$\times$480 pixels) and a \textit{FLIR} RGB frame camera (Chameleon3 USB3, 2048$\times$1536 pixels, 55FPS, lens of 4.5mm/F1.95), mounted side-by-side.
We calibrated the hybrid system using a blinking checkerboard video and computed camera intrinsic and extrinsic matrices following \cite{heikkila1997four, zhang2000flexible}.
We synchronised the two cameras by sending an external signal from the frame camera to trigger timestamped zero magnitude events in the event camera.

We obtained an HDR reference image for quantitative evaluation of a sequence via traditional multi-exposure image fusion followed by an image warp to register the reference image with each frame.
The scene in the proposed dataset is chosen to be static and far away from the camera, so that SURF feature matching \cite{bay2006surf} and homography estimation are sufficient for the image registration.

We also provide an artificial HDR (AHDR) dataset that was generated by simulating a low dynamic range (LDR) camera by applying an artificial camera response function and using the original images as HDR references.
We synthesised LDR images in this manner to provide additional data to verify the performance of our algorithm.

\section{Experiments}
\label{sec:results}
\begin{figure*}
	\newcommand{\colwidth}{3.3cm}
	\renewcommand{\tabcolsep}{0.4mm}
	\centering
	\resizebox{1\textwidth}{!}{
		\begin{tabular}
			{
				>{\centering\arraybackslash}m{4mm} >{\centering\arraybackslash}m{\colwidth} |
				>{\centering\arraybackslash}m{\colwidth} >{\centering\arraybackslash}m{\colwidth}
				>{\centering\arraybackslash}m{\colwidth} | >{\centering\arraybackslash}m{\colwidth}}
			\rotatebox{90}{\textbf{HDR} \texttt{Trees}}
			&
			\includegraphics[width=\linewidth]{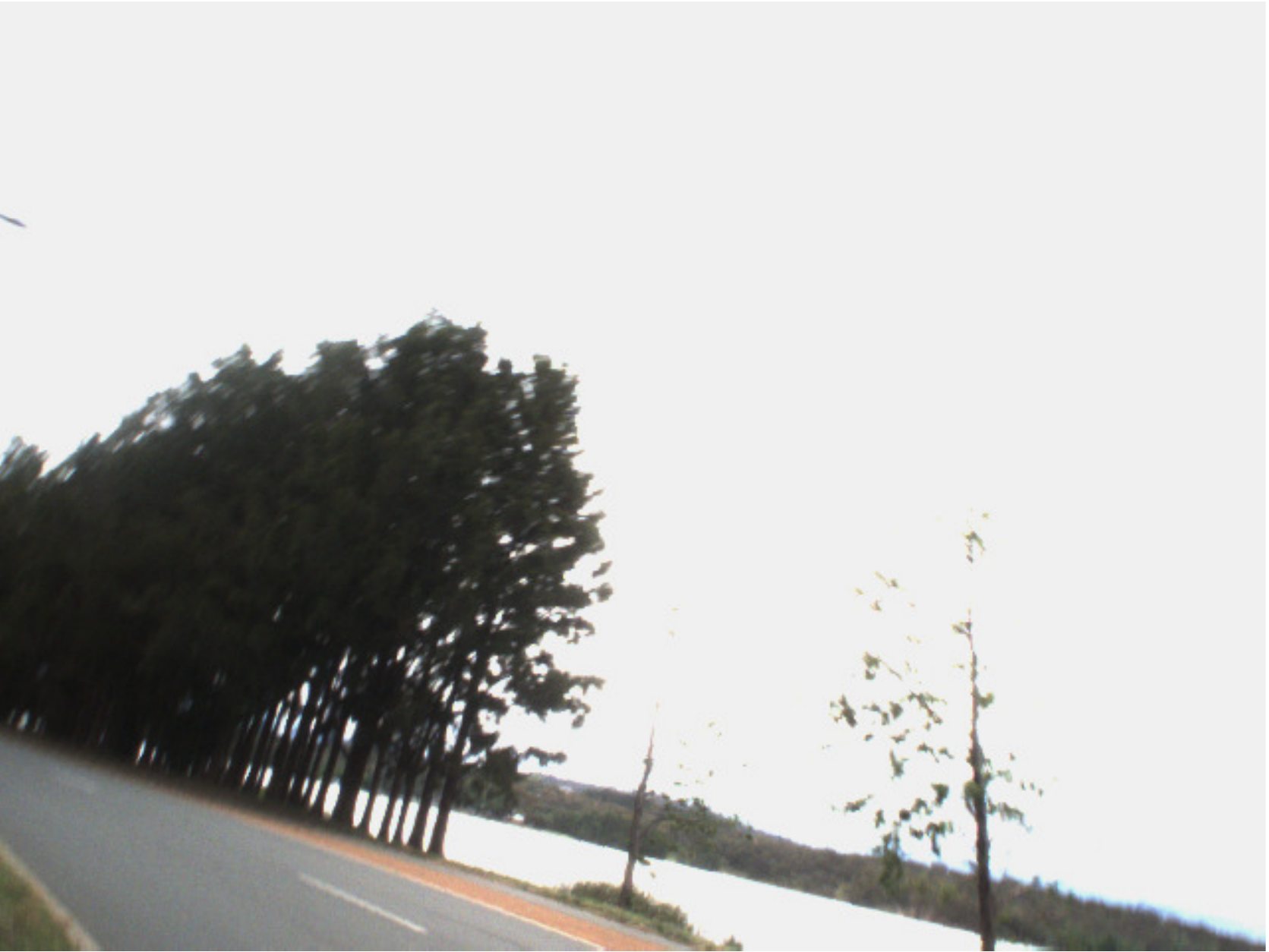}
			&
			\includegraphics[width=\linewidth]{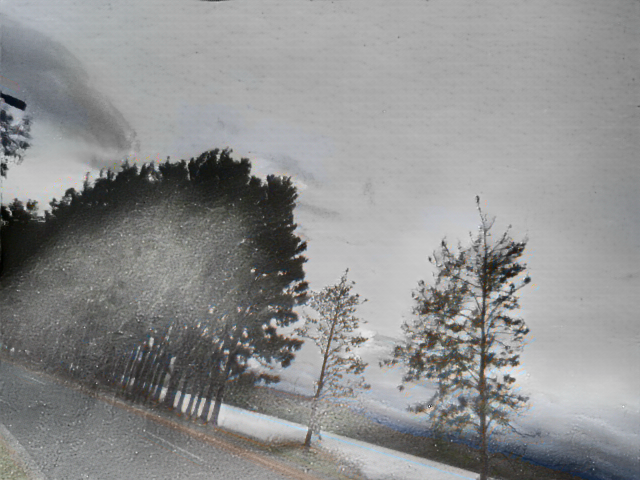}
			&
			\includegraphics[width=\linewidth]{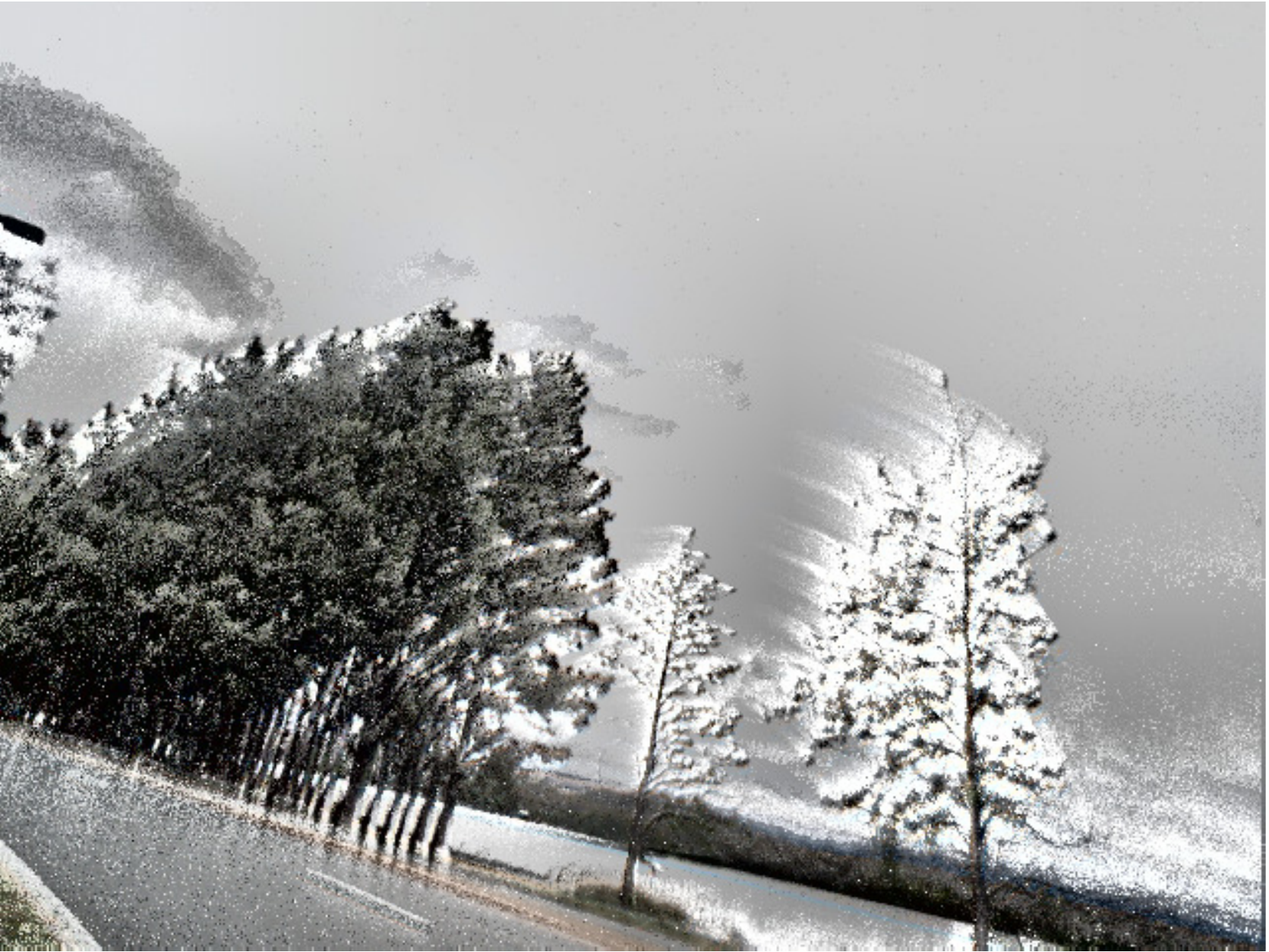}
			&
			\includegraphics[width=\linewidth]{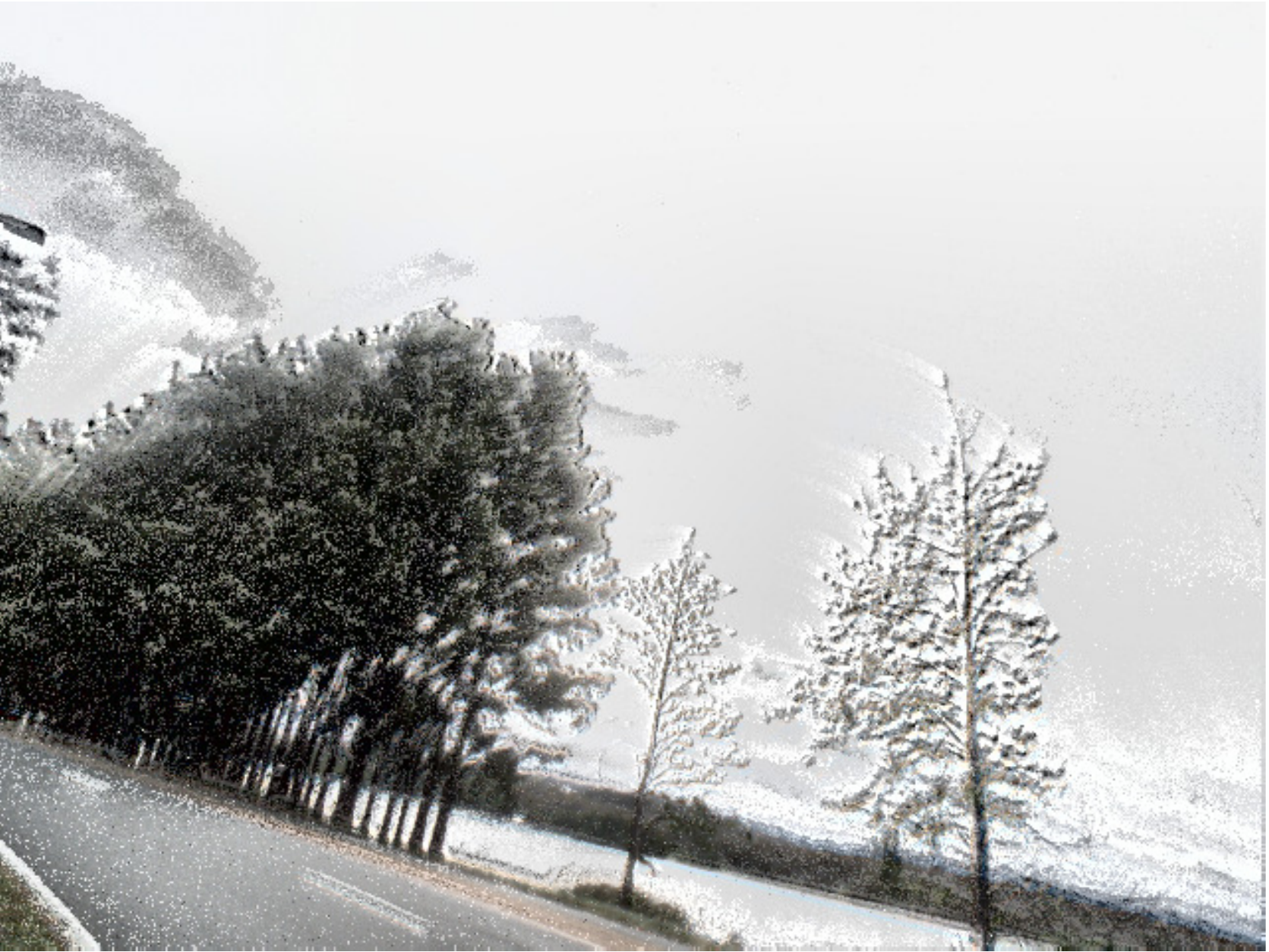}
			&
			\includegraphics[width=\linewidth]{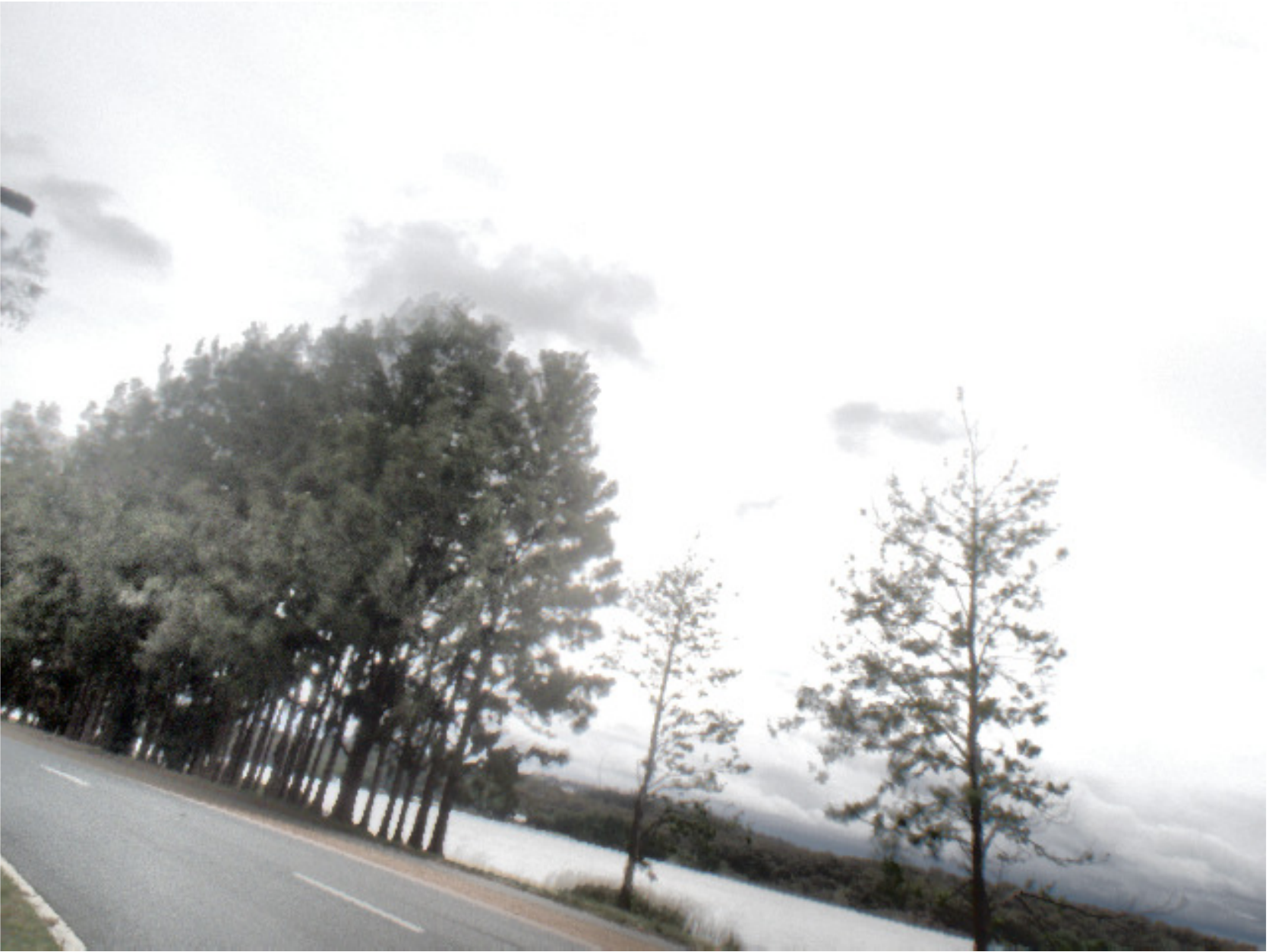}
			\\
			\midrule
			\rotatebox{90}{\textbf{AHDR} \texttt{Mountain}}
			&
			\includegraphics[width=\linewidth]{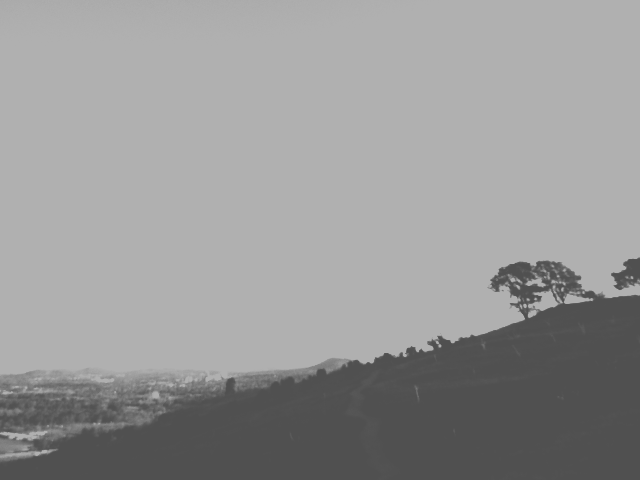}
			&
			\includegraphics[width=\linewidth]{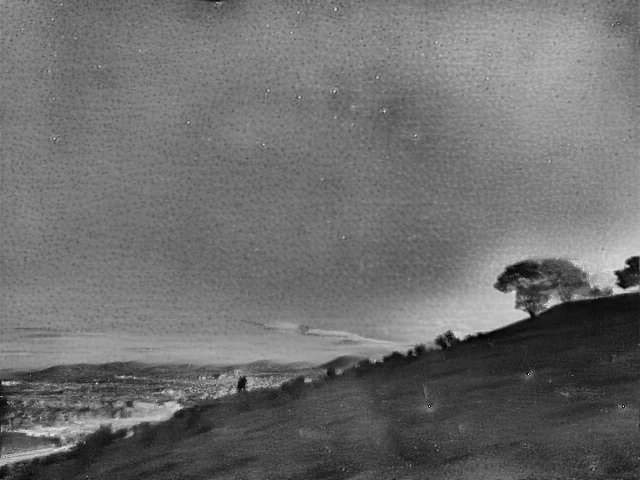}
			&
			\includegraphics[width=\linewidth]{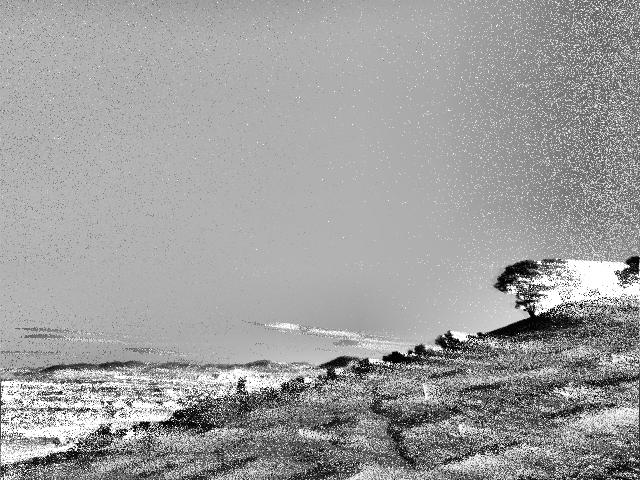}
			&
			\includegraphics[width=\linewidth]{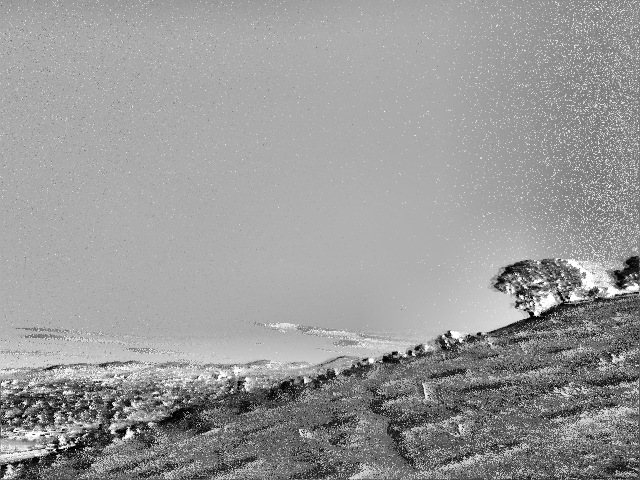}
			&
			\includegraphics[width=\linewidth]{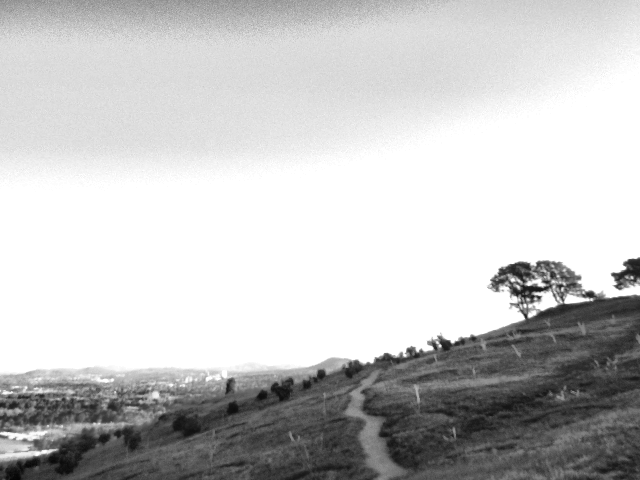}
			\\		
			& LDR input image
			& E2VID \cite{Rebecq20pami}
			& CF \cite{Scheerlinck18accv}
			& \textbf{AKF (Ours)}
			& Reference Image
		\end{tabular}
	}
	\caption{
		Typical results from the proposed HDR and AHDR dataset.
		Our HDR dataset includes referenced HDR images generated by fusing several images of various exposures.
		Our AHDR dataset is simulated by saturating the values of well-exposed real images, taking out most of the details.
		The original images are used as HDR references.
		E2VID \cite{Rebecq20pami} uses events only.
		The input images used in the CF \cite{Scheerlinck18accv} and AKF are low dynamic range.
		CF \cite{Scheerlinck18accv} leads to shadows on moving object edges.
		E2VID~\cite{Rebecq20pami} performs poorly on the dark trees in the HDR dataset and the road/sky in the AHDR dataset.
		Our AKF correctly computes the underexposed and overexposed trees in the HDR dataset and reconstructs the mountain road clearly in the artificially saturated regions.
	}
	\label{fig:hdr}
\end{figure*}

We compared our proposed Asynchronous Kalman Filter (AKF) with three state-of-the-art event-based video reconstruction methods:
E2VID \cite{Rebecq20pami} and ECNN \cite{Stoffregen20eccv} are neural networks that use only events to reconstruct video, while CF \cite{Scheerlinck18accv} is a filter-based method that combines events and frames.
In Fig.~\ref{fig:davis}, we evaluate these methods on some challenging sequences from the popular open-source event camera datasets ACD~\cite{Scheerlinck18accv}, CED~\cite{scheerlinck2019ced} and IJRR~\cite{Mueggler17ijrr}.
We also evaluate these methods on the proposed HDR and AHDR dataset in Fig.~\ref{fig:hdr} and Table~\ref{tab:hdr}.

\noindent\textbf{Evaluation:}
We quantitatively evaluated image reconstruction quality
with the HDR reference in the proposed dataset using the following metrics:
Mean squared error (MSE),
structural similarity Index Measure (SSIM) \cite{Wang04tip},
and Q-score~\cite{Narwaria15jei}.
SSIM measures the structural similarity between the reconstructions and references.
Q-score is a metric tailored to HDR full-reference evaluation.
All metrics are computed on the un-altered reconstruction and raw HDR intensities.

\begin{table*}[t]
	\centering
	\caption{Comparison of state-of-the-art event-based video reconstruction methods E2VID \cite{Rebecq20pami}, ECNN \cite{Stoffregen20eccv} and CF \cite{Scheerlinck18accv} on the proposed HDR and AHDR dataset.
		Metrics are evaluated over the full dataset of 9 sequences. 	
		Our AKF outperforms the compared methods on all metrics.
		Detailed evaluation on each sequence can be found in the supplementary material. Higher SSIM and Q-score and lower MSE indicate better performance.
	}
	\label{tab:hdr}
	\resizebox{0.95\textwidth}{!}{
		\begin{tabular}{ l | c c c c | c c c c | c c c c }
			Metrics &  \multicolumn{4}{c|}{MSE ($\times 10^{-2}$) $\downarrow$}  &  \multicolumn{4}{c|}{SSIM~\cite{Wang04tip} $\uparrow$}
			&    \multicolumn{4}{c}{Q-score~\cite{Narwaria15jei} $\uparrow$}
			\\
			\midrule
			Methods & E2VID & ECNN  & CF & AKF (ours) & E2VID & ECNN  & CF & AKF (ours) & E2VID & ECNN  & CF & AKF (ours)
			\\
			\midrule
			\midrule
			HDR & $7.76 $ & $11.43 $ & $6.22 $  & $\textbf{1.71} $  & $0.616 $ & $0.31 $ & $0.66  $ & $\textbf{0.89} $  & $4.32 $ & $3.41 $ & $3.01 $ & $\textbf{4.83} $
			
			\\
			\midrule
			AHDR & $11.56 $ & $21.23 $ & $5.28 $  & $\textbf{4.18} $ & $0.50 $ & $0.04 $ & $0.62  $ & $\textbf{0.75}$  & $5.24 $ & $3.36 $ & $4.78 $ & $\textbf{5.54} $
			\\
			\bottomrule[\heavyrulewidth]
		\end{tabular}
	}
\end{table*}


\noindent\textbf{Main Results:}
The open-source event camera datasets ACD~\cite{Scheerlinck18accv}, CED~\cite{scheerlinck2019ced} and IJRR~\cite{Mueggler17ijrr} are popularly used in several event-based video reconstruction works.
Without HDR references, we only visually evaluate on the challenging HDR scenes from these datasets in Fig.~\ref{fig:front page} and \ref{fig:davis}.
\texttt{Night drive} investigates extreme low-light, fast-speed, night driving scenario with blurry and underexposed/overexposed DAVIS frames.
\texttt{Shadow} evaluates the scenario of static background, dynamic foreground objects with overexposed region.
\texttt{Outdoor running} evaluates the outdoor overexposed scene with event camera noise.
Both AKF and E2VID~\cite{Rebecq20pami} are able to capture HDR objects (\eg, right turn sign in \texttt{Night drive}),
but E2VID~\cite{Rebecq20pami} fails to capture the background in \texttt{Shadow} because the stationary event camera provides no information about the static background.
In \texttt{Outdoor running}, it is clear that E2VID~\cite{Rebecq20pami} is unable to reproduce the correct high dynamic range intensity between the dark road and bright left building and sky background.
Our AKF algorithm is able to resolve distant buildings despite the fact that they are too bright and washed out in the LDR DAVIS frame.
The cutoff frequency of CF~\cite{Scheerlinck18accv}, which corresponds to the Kalman gain of our AKF is a single constant value for all pixels.
This causes CF~\cite{Scheerlinck18accv} to exhibits `shadowing effect' on object edges (on the trailing edge of road sign and buildings).
AKF overcomes the `shadowing effect' by dynamically adjusting the per-pixel Kalman gain based on our uncertainty model.
Our frame augmentation also sharpens the blurry DAVIS frame and reduces temporal mismatch between the high data rate events and the low data rate frames.
AKF reconstructs the sharpest and most detailed HDR objects in all challenging scenes.

Table~\ref{tab:hdr} shows that our AKF outperforms other methods on the proposed HDR/AHDR dataset on MSE, SSIM and Q-score.
Unsurprisingly, our AKF outperforms E2VID~\cite{Rebecq20pami} and ECNN~\cite{Stoffregen20eccv} since it utilises frame information in addition to events.
CF~\cite{Scheerlinck18accv} performs worse compared to E2VID~\cite{Rebecq20pami} and ECNN~\cite{Stoffregen20eccv} in some cases despite utilising frame information in addition to events.
AKF outperforms state-of-the-art methods in the absolute intensity error MSE with a significant reduction of 48\% and improve the image similarity metrics SSIM and Q-score by 11\% on average.
The performance demonstrates the importance of taking into account frame and event noise and preprocessing frame inputs compared to CF~\cite{Scheerlinck18accv}.

Fig.~\ref{fig:hdr} shows qualitative samples of input, reconstructed and reference images from the proposed HDR/AHDR dataset.
In the first row of Fig.~\ref{fig:hdr}, the proposed HDR dataset
\texttt{Trees} includes some underexposed trees (left-hand side) and two overexposed trees (right-hand side).
In the second row,
our AHDR sequence \texttt{Mountain} is artificially saturated (pixel values higher than 160 or lower than 100 of an 8-bit image), removing most of the detail.
E2VID~\cite{Rebecq20pami} reconstructs the two right-hand trees correctly, although the relative intensity of the tree is too dark.
E2VID~\cite{Rebecq20pami} also performs poorly in the dark area in \texttt{Trees} on the bottom left corner and skies/road in \texttt{Mountain} where it lacks events.
CF~\cite{Scheerlinck18accv} exhibits `shadowing effect' on object edges (trees and mountain road), which is significantly reduced in AKF by dynamically adjusting the per-pixel Kalman gain according to events and frame uncertainty model.

\section{Conclusion}
In this paper, we introduced an asynchronous Kalman-Bucy filter to reconstruct HDR videos from LDR frames and event data for fast-motion and blurry scenes.
The Kalman gain is estimated pixel-by-pixel based on a unifying \eventframe uncertainty model over time.
In addition, we proposed a novel frame augmentation algorithm that can also be widely applied to many existing event-based applications.
To target HDR reconstruction, we presented a real-world, hybrid \eventframe dataset captured on registered frame and event cameras.
We believe our asynchronous Kalman filter has practical applications for video acquisition in HDR scenarios using the extended power of event cameras in addition to conventional frame-based cameras.
\clearpage

{\small
	\bibliographystyle{ieee_fullname}
	\bibliography{template_arxiv,ced}
}

\end{document}